\documentclass{article}

\usepackage[nonatbib,final]{neurips_2019_ml4ps}

\usepackage[utf8]{inputenc} 
\usepackage[T1]{fontenc}    
\usepackage{hyperref}       
\usepackage{url}            
\usepackage{booktabs}       
\usepackage{amsfonts}       
\usepackage{nicefrac}       
\usepackage{microtype}      

\usepackage[numbers,square]{natbib}
\usepackage{graphicx}
\usepackage{bm}
\usepackage{listings}
\usepackage{amsmath}
\usepackage{caption}
\usepackage[ruled,noend]{algorithm2e}
\title{Sim-to-Real Domain Adaptation For High Energy Physics}

\author{Marouen~Baalouch\textsuperscript{1}
\And
Jean-Philippe Poli\textsuperscript{1}
\And
Maxime Defurne\textsuperscript{2}
\And
Noëlie Cherrier\textsuperscript{2}
\and
\textsuperscript{1} CEA, LIST, 91191, Gif-sur-Yvette cedex, France.\\
\textsuperscript{2} Irfu, CEA, Universit\'e Paris-Saclay, 91191, Gif-sur-Yvette cedex, France.
}

\begin{document}

\maketitle

\begin{abstract}
Particle physics or High Energy Physics (HEP) studies the elementary constituents of matter and their interactions with each other. Machine Learning (ML) has played an important role in HEP analysis and has proven extremely successful in this area. Usually, the ML algorithms are trained on numerical simulations of the experimental setup and then applied to the real experimental data. However, any discrepancy between the simulation and real data may lead to dramatic consequences concerning the performances of the algorithm on real data. In this paper, we present an application of domain adaptation using a Domain Adversarial Neural Network trained on public HEP data. We demonstrate the success of this approach to achieve sim-to-real transfer and ensure the consistency of the ML algorithms performances on real and simulated HEP datasets.  
\end{abstract}

\section{Introduction}


Elementary particles and the way they interact with each other are studied in particle collider facilities such as the Large Hadron Collider (LHC)~\cite{Evans_2008}. Out of the collision between two particles, new particles will emerge and interact with the detectors placed around the collision site. These detectors provide information such as the particle position, the arrival time and its energy deposit inside the detector which allow to reconstruct the particle trajectory and identify it. Between the produced particles, various correlations and quantities are computed to classify the collision. Classically based on advanced statistics, physicists study patterns and/or compare the results of this classification to Standard Model predictions, hoping to find a deviation hinting ``New Physics''. 

However, in the recent years, Machine Leaning (ML) brought new levels of performance in the classification exercise. In real experimental data, the labels are rarely available and, as a consequence, ML classifiers are usually trained on numerical simulations. The simulation is divided into two main parts: simulation of the high-energy physics processes, $e.g.$, collision/production of particles, and simulation of the detector responses to the produced particles. Both parts may need approximations due to the complexity of the physics processes, causing discrepancies between simulated and experimental data which could dramatically lower the classifier performances with real data.  

In this work, we are performing the classification of a LHCb public dataset~\footnote{Flavours of Physics: finding $\tau \to \mu\mu\mu$. www.kaggle.com/c/flavours-of-physics/overview} with a Domain-Adversarial Neural Network (DANN), a Transfer Learning (TL) technique, and compare it to the performances of a regular Neural Network both trained on the same simulation. 


\section{Dataset}
 \label{datasec}
 The dataset studied in this paper was produced for the LHCb experiment at CERN \cite{Alves:2008zz}. The main goal of the LHCb experiment is the search for possible contributions from physics beyond the Standard Model by performing precision measurements of {\it CP} violating observables and rare decays of hadrons containing a $b$ quark or a $c$ quark. LHCb collaboration posted a data challenge on Kaggle site, for ML application purpose, whose aim is to find charged lepton flavour violation, a possible signal of new physics. The main task of this analysis is to design a ML classifier to select the data of interest (signal) consisting of the decay of the elementary lepton particle $\tau$ to the three quasi-stable lepton particle $\mu$. The LHCb dataset available for the training consists of mixture of simulated and real data which are respectively used to describe the signal and background. The dataset feature space is composed by $58$ variables, reconstructed from the detector responses and representing physical quantities that describe $\tau$-decay. The dataset contains $67533$ examples, where $62 \%$ are signal events.      
  
 Another dataset, called control sample, was provided to evaluate the agreement of the classifier on simulated and real data. This dataset contains simulated and real events from the channel decay $D_s \to \phi \pi \to 3\mu$, that has a similar topology as the signal channel decay $\tau \to 3\mu$ and the same feature space. The simulation (source) contains $8205$ examples and the real data (target) contains $322942$ examples with weight values. The weights are estimated using $sPlot$ method \cite{splot}. The weight of each example describes a probability to be signal or background: higher (resp. lower) weights mean this event is likely to be signal (resp. background).   
 The agreement of the trained classifier is evaluated using the Kolmogorov Smirnov distance between the classifier probability output from source control (simulation) and target control (real data).
 
Since the same framework was used to collect the examples for both training and control samples, the shifts between real data and simulation for the process of interest $\tau \to 3\mu$ are expected to be similar to the shifts for the control process $D_s \to \phi \pi \to 3\mu$. The goal is then to learn simulation/real experiment discrepancies with the control process and then transfer this knowledge to the classifier searching for the $\tau \to 3\mu$ process in experimental data.

\section{Related Work}
\label{sec:relatedwork}

Transfer learning consists in reusing a pre-trained model on a new problem that is, in a certain way, more or less related to the first one. It has been carried out successfully in several fields as image recognition~\cite{Gatys_2016_CVPR, 6126344, Oquab_2014_CVPR}, sentiment classification~\cite{Glorot:2011:DAL:3104482.3104547, Ng:2015:DLE:2818346.2830593} or robotics~\cite{8460528, DBLP:journals/corr/RusuVRHPH16, DBLP:journals/corr/abs-1809-04720}. Sim-to-real transfer has been particularly studied in robotics \cite{DBLP:journals/corr/abs-1809-04720,pmlr-v87-golemo18a}: it is a sub-domain of transfer learning that aims at transferring a model trained on simulated data to real data. This is a paramount approach for HEP data. Indeed, the simulation modeling solves real-world problems safely and efficiently and allows the generation of unlimited size datasets. For ML application, learning from a simulation and applying the acquired knowledge to the real world can provide a robust ML models with less time-consuming and data search, specially for ML application related to hardware interaction. However, the simulation is not able to fully replicate all reactions in the real world, which can create a dataset shift between the two domains, simulation and reality. The application of transfer learning in physics is growing due the wide use of the computing simulation in different analysis. It has been applied as example in the modeling of inertial confinement fusion experiments~\cite{DBLP:journals/corr/abs-1812-06055}, galaxy detection~\cite{galaxy, Khan:2018opv},  or gravitational wave detection~\cite{PhysRevD.97.101501}.     

HEP experiments are growing in size and complexity and computer simulation of these experiments are essential for analyzing and interpreting experimental data. 

The dataset shift between HEP simulation and real experimental data is also present, and is mainly relying on the complexity of the physics interactions at this level of energy and scale.

\section{Sim-to-Real domain adaptation }
\label{sec:dataset}
\subsection{Motivations}
The first motivation of this work is that simulations provide a lot of labeled data whereas there are very few real labeled data. So far, the process of real data is unavoidable. Applying a classifier trained on simulations on real data decreases indeed the performance of classification, while mainly increasing the systematic uncertainty which lowers the significance of the results.

 To achieve that, characterizing the dataset shift between simulation and real experiment gives an idea about the appropriate transfer learning technique for a specific learning task. The transfer of ML algorithm trained on HEP simulation represents a task of {\it Domain Adaptation} (DA)~\cite{Ben-David2010, Patel2015VisualDA, Pan:2010:STL:1850483.1850545}. In this case of DA, the source with labels is provided by the HEP simulation and the target without label by the events from the detector response.
 We aim in this ML application study at selecting an event of interest by physics process classification, with an ML model trained on simulation and targeted to real data. In this HEP study context, the dataset shift can be categorized under three types~\cite{article}:
 
 \begin{itemize}
     \item the prior probability shift occurs when the estimation of the total rate of background or data of interest in the simulation is not the same as in the reality; 
     \item the covariate shift occurs when the geometry of the physics process, for instance the angular or energy distribution of the produced particles, is approximate or when the performance of the detector is not ideally realistic in the simulation framework. However the rate of data of interest with respect to the background is correctly defined; 
     \item the concept shift occurs when the simulation are not considering all the categories of physics processes in the real data, thus are counted as a form of signal or background.
 \end{itemize}
 
  In the case of fully unlabeled target, the prior probability and covariate shift can be corrected. However, the concept shift requires labeled target data since this type of shift is related to data drift, where classifiers are deployed in non-stationary environments~\cite{Widmer1996}.
 
 The studied dataset, described in section~\ref{datasec}, is expected to contain a covariate shift between the source and target domain. In this case, the two related domains, described by the same input and output space $\mathcal{X}$ and $\mathcal{Y}$, have an equivalent posterior distribution $p(x|y)$, but different probability input distributions $p(x)$.
 
 \subsection{Domain Adaptation with Adversarial Network}
 Domain adaptation is achieved by training a model on labeled data $S$ from a source domain $\mathcal{D}_{\rm S}$ while minimizing test error on a target domain $\mathcal{D}_{\rm T}$, for which no labels in the target data $T$ are available at training time. Driven by a simple assumption, the source risk $R_{\mathcal{D}_{\rm S}}$ is expected to be a good indicator of the target risk $R_{\mathcal{D}_{\rm T}}$ when both distributions are similar. This approach is validated by the  theory obtained by Ben David et al.~\cite{Ben-David2010, NIPS2006_2983}, proving that for an effective domain transfer to be achieved, predictions must be made based on a data representation that cannot discriminate between the source and target domains.  
 

 
Several DA approaches implement this idea of domain similarity space (feature alignment) in the context of neural network architectures, by minimizing statistical distance between distributions~\cite{pmlr-v37-long15, DBLP:journals/corr/SunS16a} or by adversarial domain alignment~\cite{2014arXiv1409.7495G, DBLP:journals/corr/TzengHDS15}.  

 A Domain-Adversarial Neural Network (DANN)~\cite{2014arXiv1409.7495G} represents an appropriate approach to learn a HEP classifier that can generalize well from simulation to real experimental context. It consists in a feed-forward network with added standard layers and a gradient reversal layer.
 
 This method is easy to implement since DANN can be trained with standard backpropagation and stochastic gradient descent (available in every deep learning libraries), less time-consuming with respect to domain distribution alignment with statistical distance minimization and can easy handle weighted examples, which motivated the use of this approach in the current HEP data analysis.    
  The feature alignment can then be performed with the control process: in the DANN, the network hidden layer working adversarially towards output connections predicting domain membership learns from control process examples instead of the process-of-interest examples. 
 
\section{Experiment and results}
\label{sec:exp}
 To demonstrate the performance of DANN on LHCb dataset, we trained a benchmark classifier based on standard feedforward neural network, containing one input layer with the size of the input space, one hidden layer with 100 neurons and an output layer with two neurons. In this architecture, we use hyperbolic tangent as activation function and the cross-entropy as a cost function. In the other hand, the DANN model is designed with the same architecture of the benchmark model but in addition to the output layer, the hidden layer is linked to a gradient reversal layer with two output neurons used to classify the domain (simulation or real data) using control channel examples. Both ML architectures are trained using standard backpropagation and stochastic gradient-based optimizer {\it Adam}. The standard NN is trained only on the training dataset and the DANN is trained on training (class) and control (domain) data. 

\begin{figure}[!h]
  \centering
     \includegraphics[width=0.85\linewidth]{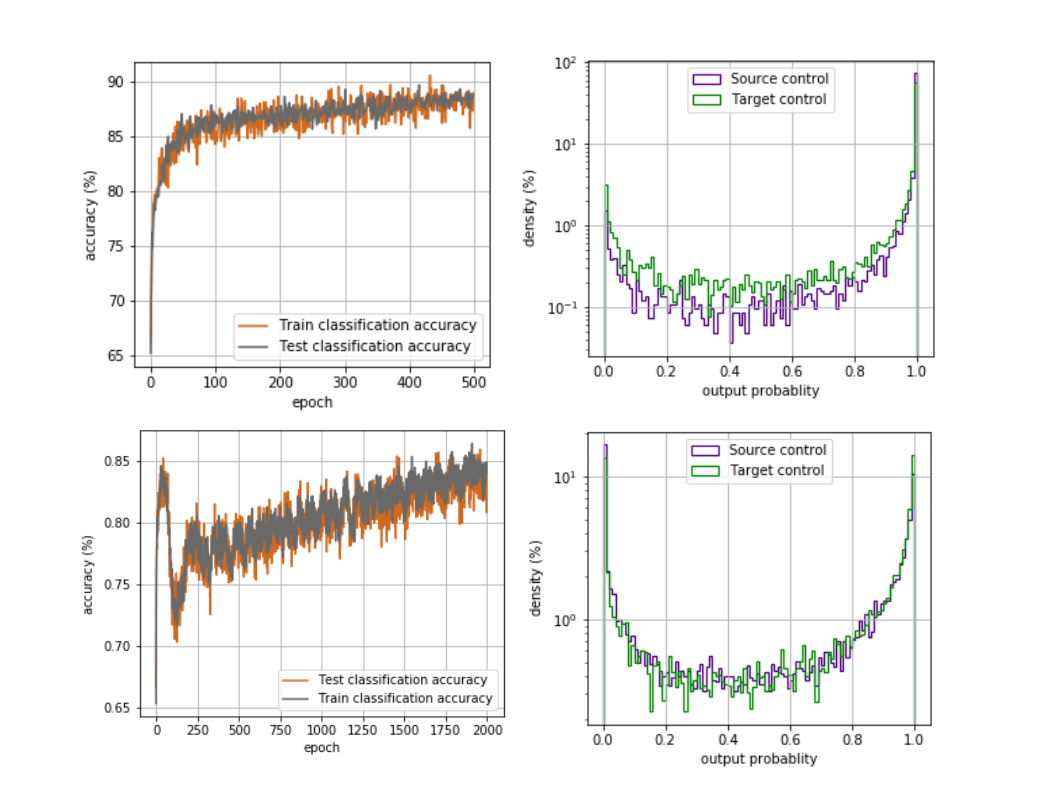}
  \caption{Left: Accuracy values along NN (top) and DANN (bottom) models optimization. Right: Output prediction probability using trained NN (top) and DANN (bottom). The overlap of source and target curve for DANN demonstrates the effectiveness of the domain adaptation.}
  \label{fig:loss}
\end{figure}

For both classification tasks, with and without adversarial network, the datasets are split into two subsets, one for the training, containing $70 \%$ of the initial data, and one for the evaluation, named testing dataset. During the optimization, the training dataset is divided into $22$ batches with size of $3000$ examples. Figure~\ref{fig:loss} shows the evolution of the accuracy for both models function of the epoch iteration. Simply based on accuracy and convergence speed, the NN classifier outperforms the DANN classifier, with an accuracy of $88.1 \%$ against $84.0 \%$ on simulation as shown by Figure~\ref{fig:loss}. 

To ensure the conservation of these performances with real data, a Kolmogorov Smirnov distance between the output probability distributions of the classifiers on source and target for the process of interest (displayed in Figure~\ref{fig:loss}) must be found below 0.09: for the standard NN, the distance is $0.19$ and for the DANN, it is $0.06$. The performances of the DANN-classifier are therefore conserved between simulation and experimental data while they are not for the standard NN.   

\section{Conclusion}
Since HEP simulation is never exactly representative of experiments, a robust characterization of the ML classifier performances on unlabeled real data is hard to achieve. Such a characterization being mandatory for publication, domain adaptation from HEP simulation to real HEP experimental data is essential. 

 In this paper, we categorized the possible sources of dataset shifts between simulation and real data in a realistic HEP data challenge. Domain adaptation is achieved by training a DANN: trained with a control physics process with similar geometry as the process-of-interest, its adversarial part ensures that the resulting ML classifier is as little biased as possible by shifts between simulation and reality. A Kolmogorov-Smirnov test between the output probability distribution of the classifier on simulation and on real data for the process-of-interest confirms that the classifier is unbiased and, therefore, its performances are identical on simulation and real data unlike a standard NN-classifier. In other words, any signal found in the experimental data by the DANN-classifier would be directly validated as a discovery of physics beyond the standard model.




\bibliographystyle{unsrtnat} 
\bibliography{main}

\end{document}